\documentclass[english,letterpaper,journal,compsoc]{IEEEtran}
\usepackage[T1]{fontenc}
\usepackage[utf8]{inputenc}
\usepackage{amsmath}
\usepackage{amssymb}
\usepackage{esint}

\makeatletter

\providecommand{\tabularnewline}{\\}

\makeatother

\usepackage{babel}
\begin{document}

\title{Mixture Gaussian Process Conditional Heteroscedasticity}

\author{Emmanouil A. Platanios and Sotirios P. Chatzis%
\thanks{E. A. Platanios is currently with Imperial College London. He conducted
this research as part of an internship with the Department of Electrical
Engineering, Computer Engineering, and Informatics, Cyprus University
of Technology. S. P. Chatzis is with the Department of Electrical
Engineering, Computer Engineering, and Informatics, Cyprus University
of Technology.%
}}
\maketitle
\begin{abstract}
Generalized autoregressive conditional heteroscedasticity (GARCH)
models have long been considered as one of the most successful families
of approaches for volatility modeling in financial return series.
In this paper, we propose an alternative approach based on methodologies
widely used in the field of statistical machine learning. Specifically,
we propose a novel nonparametric Bayesian mixture of Gaussian process
regression models, each component of which models the noise variance
process that contaminates the observed data as a separate latent Gaussian
process driven by the observed data. This way, we essentially obtain
a mixture Gaussian process conditional heteroscedasticity (MGPCH)
model for volatility modeling in financial return series. We impose
a nonparametric prior with power-law nature over the distribution
of the model mixture components, namely the Pitman-Yor process prior,
to allow for better capturing modeled data distributions with heavy
tails and skewness. Finally, we provide a copula-based approach for
obtaining a predictive posterior for the covariances over the asset
returns modeled by means of a postulated MGPCH model. We evaluate
the efficacy of our approach in a number of benchmark scenarios, and
compare its performance to state-of-the-art methodologies.\end{abstract}
\begin{IEEEkeywords}
Gaussian process, Pitman-Yor process, mixture model, conditional heteroscedasticity,
copula, volatility modeling.
\end{IEEEkeywords}

\section{Introduction}

Statistical modeling of asset values in financial markets requires
taking into account the tendency of assets towards asymmetric temporal
dependence \cite{chollette}. Besides, the data generation processes
of the returns of financial market indexes may be non-linear, non-stationary
and/or heavy-tailed, while the marginal distributions may be asymmetric,
leptokurtic and/or show conditional heteroscedasticity. Hence, there
is a need to construct flexible models capable of incorporating these
features. The generalized autoregressive conditional heteroscedasticity
(GARCH) family of models has been used to address conditional heteroscedasticity
and excess kurtosis (see, e.g., \cite{engle,bolerslev}). 

The time-dependent variance in series of returns on prices, also known
as volatility, is of particular interest in finance, as it impacts
the pricing of financial instruments, and it is a key concept in market
regulation. GARCH approaches are commonly employed in modeling financial
return series that exhibit time-varying volatility clustering, i.e.
periods of swings followed by periods of relative calm, and have been
shown to yield excellent performance in these applications, consistently
defining the state-of-the-art in the field in the last decade. GARCH
models represent the variance by a function of the past squared returns
and the past variances, which facilitates model estimation and computation
of the prediction errors.

Gaussian process (GP) models comprise one of the most popular Bayesian
methods in the field of machine learning for regression, function
approximation, and predictive density estimation \cite{tnn1}. Despite
their significant flexibility and success in many application domains,
GPs do also suffer from several limitations. In particular, GP models
are faced with difficulties when dealing with tasks entailing non-stationary
covariance functions, multi-modal output, or discontinuities. Several
approaches that entail using ensembles of fractional GP models defined
on subsets of the input space have been proposed as a means of resolving
these issues (see, e.g., \cite{igp,igp2,igp3}). 

In this work, we propose a novel GP-based approach for volatility
modeling in financial time series (return) data. Our proposed approach
provides a viable alternative to GARCH models, that allows for effectively
capturing the clustering effects in the variability or volatility.
Our approach is based on the introduction of a novel nonparametric
Bayesian mixture model, the component distributions of which constitute
GP regression models; the noise variance processes of the model component
GPs are considered as input-dependent latent variable processes which
are also modeled by imposition of appropriate GP priors. This way,
our novel approach allows for learning both the observation-dependent
nature of asset volatility, as well as the underlying volatility clustering
mechanism, modeled as a latent model component switching procedure.
We dub our approach the mixture Gaussian process conditional heteroscedasticity
(MGPCH) model.

Nonparametric Bayesian modeling techniques, especially Dirichlet process
mixture (DPM) models, have become very popular in statistics over
the last few years, for performing nonparametric density estimation
\cite{dpm2,dpm3,dpm4}. Briefly, a realization of a DPM can be seen
as an infinite mixture of distributions with given parametric shape
(e.g., Gaussian). An interesting alternative to the Dirichlet process
prior for nonparametric Bayesian modeling is the Pitman-Yor process
prior \cite{py}. Pitman-Yor processes produce a small number of sparsely
populated clusters and a large number of highly populated clusters
\cite{teh}. Indeed, the Pitman-Yor process prior can be viewed as
a generalization of the Dirichlet process prior, and reduces to it
for a specific selection of its parameter values. Consequently, the
Pitman-Yor process turns out to be more promising as a means of modeling
complex real-life datasets that usually comprise a high number of
clusters which comprise only few data points, and a low number of
clusters which are highly frequent, thus dominating the entire population.

Inspired by these advances, the component switching mechanism of our
model is obtained by means of a Pitman-Yor process prior imposed over
the component GP latent allocation variables of our model. We derive
a computationally efficient inference algorithm for our model based
on the variational Bayesian framework, and obtain the predictive density
of our model using an approximation technique. We examine the efficacy
of our approach considering volatility prediction in a number of financial
return series. 

The remainder of this paper is organized as follows: In Section 2,
we provide a brief presentation of the theoretical background of the
proposed method. Initially, we present the Pitman-Yor process and
its function as a prior in nonparametric Bayesian models; further,
we provide a brief summary of Gaussian process regression. In Section
3, we introduce the proposed mixture Gaussian process conditional
heteroscedasticity (MGPCH) model, and derive efficient model inference
algorithms based on the variational Bayesian framework. We also propose
a copula-based method for learning the interdependencies between the
returns of multiple assets jointly modeled by means of an MGPCH model.
In Section 4, we conduct the experimental evaluation of our proposed
model, considering a number of applications dealing with volatility
modeling in financial return series. In the final section, we summarize
and discuss our results.

\section{Preliminaries}

\subsection{The Pitman-Yor process}

Dirichlet process models were first introduced by Ferguson \cite{ferguson-dp}.
A DP is characterized by a base distribution $G_{0}$ and a positive
scalar $\alpha$, usually referred to as the innovation parameter,
and is denoted as $\mathrm{DP}(\alpha,G_{0})$. Essentially, a DP
is a distribution placed over a distribution. Let us suppose we randomly
draw a sample distribution $G$ from a DP, and, subsequently, we independently
draw $M$ random variables $\{\Theta_{m}^{*}\}_{m=1}^{M}$ from $G$:
\begin{equation}
G|\alpha,G_{0}\sim\mathrm{DP}(\alpha,G_{0})
\end{equation}
 
\begin{equation}
\Theta_{m}^{*}|G\sim G,\quad m=1,\dots M
\end{equation}
Integrating out $G$, the joint distribution of the variables $\{\Theta_{m}^{*}\}_{m=1}^{M}$
can be shown to exhibit a clustering effect. Specifically, given the
first $M-1$ samples of $G$, $\{\Theta_{m}^{*}\}_{m=1}^{M-1}$, it
can be shown that a new sample $\Theta_{M}^{*}$ is either (a) drawn
from the base distribution $G_{0}$ with probability $\frac{\alpha}{\alpha+M-1}$,
or (b) is selected from the existing draws, according to a multinomial
allocation, with probabilities proportional to the number of the previous
draws with the same allocation \cite{polya-urn}. Let $\{\Theta_{c}\}_{c=1}^{C}$
be the set of distinct values taken by the variables $\{\Theta_{m}^{*}\}_{m=1}^{M-1}$.
Denoting as $\nu_{c}^{M-1}$ the number of values in $\{\Theta_{m}^{*}\}_{m=1}^{M-1}$
that equal to $\Theta_{c}$, the distribution of $\Theta_{M}^{*}$
given $\{\Theta_{m}^{*}\}_{m=1}^{M-1}$ can be shown to be of the
form \cite{polya-urn} 
\begin{equation}
\begin{aligned}p(\Theta_{M}^{*}|\{\Theta_{m}^{*}\}_{m=1}^{M-1},\alpha,G_{0})= & \frac{\alpha}{\alpha+M-1}G_{0}\\
 & +\sum_{c=1}^{C}\frac{\nu_{c}^{M-1}}{\alpha+M-1}\delta_{\Theta_{c}}
\end{aligned}
\end{equation}
where $\delta_{\Theta_{c}}$ denotes the distribution concentrated
at a single point $\Theta_{c}$. These results illustrate two key
properties of the DP scheme. First, the innovation parameter $\alpha$
plays a key-role in determining the number of distinct parameter values.
A larger $\alpha$ induces a higher tendency of drawing new parameters
from the base distribution $G_{0}$; indeed, as $\alpha\rightarrow\infty$
we get $G\rightarrow G_{0}$. On the contrary, as $\alpha\rightarrow0$
all $\{\Theta_{m}^{*}\}_{m=1}^{M}$ tend to cluster to a single random
variable. Second, the more often a parameter is shared, the more likely
it will be shared in the future.

The Pitman-Yor process (PYP) \cite{py} functions similar to the Dirichlet
process. Let us suppose we randomly draw a sample distribution $G$
from a PYP, and, subsequently, we independently draw $M$ random variables
$\{\Theta_{m}^{*}\}_{m=1}^{M}$ from $G$: 
\begin{equation}
G|\delta,\alpha,G_{0}\sim\mathrm{PY}(\delta,\alpha,G_{0})
\end{equation}
 with 
\begin{equation}
\Theta_{m}^{*}|G\sim G,\quad m=1,\dots M\;
\end{equation}
where $\delta\in[0,1)$ is the discount parameter of the Pitman-Yor
process, $\alpha>-\delta$ is its innovation parameter, and $G_{0}$
the base distribution. Integrating out $G$, similar to Eq. (3), we
now yield 
\begin{equation}
\begin{aligned}p(\Theta_{M}^{*}|\{\Theta_{m}^{*}\}_{m=1}^{M-1},\delta,\alpha,G_{0})= & \frac{\alpha+\delta C}{\alpha+M-1}G_{0}\\
 & +\sum_{c=1}^{C}\frac{\nu_{c}^{M-1}-\delta}{\alpha+M-1}\delta_{\Theta_{c}}
\end{aligned}
\end{equation}

As we observe, the PYP yields an expression for $p(\Theta_{M}^{*}|\{\Theta_{m}^{*}\}_{m=1}^{M-1},G_{0})$
quite similar to that of the DP, also possessing the rich-gets-richer
clustering property, i.e., the more samples have been assigned to
a draw from $G_{0}$, the more likely subsequent samples will be assigned
to the same draw. Further, the more we draw from $G_{0}$, the more
likely a new sample will again be assigned to a new draw from $G_{0}$.
These two effects together produce a \emph{power-law distribution}
where many unique $\Theta_{m}^{*}$ values are observed, most of them
rarely \cite{py}, thus allowing for better modeling observations
with heavy-tailed distributions. In particular, for $\delta>0$, the
number of unique values scales as $\mathcal{O}(\alpha M^{\delta})$,
where $M$ is the total number of draws. Note also that, for $\delta=0$,
the Pitman-Yor process reduces to the Dirichlet process, in which
case the number of unique values grows more slowly at $\mathcal{O}(\alpha\mathrm{log}M)$
\cite{teh}.

A characterization of the (unconditional) distribution of the random
variable $G$ drawn from a PYP, $\mathrm{PY}(\delta,\alpha,G_{0})$,
is provided by the stick-breaking construction of Sethuraman \cite{stick-break}.
Consider two infinite collections of independent random variables
$\boldsymbol{v}={(v_{c})}_{c=1}^{\infty}$, $\{\Theta_{c}\}_{c=1}^{\infty}$,
where the $v_{c}$ are drawn from a Beta distribution, and the $\Theta_{c}$
are independently drawn from the base distribution $G_{0}$. The stick-breaking
representation of $G$ is then given by \cite{teh} 
\begin{equation}
G=\sum_{c=1}^{\infty}\varpi_{c}(\boldsymbol{v})\delta_{\Theta_{c}}
\end{equation}
 where
\begin{equation}
p(v_{c})=\mathrm{Beta}(1-\delta,\alpha+\delta c)
\end{equation}
\begin{equation}
\varpi_{c}(\boldsymbol{v})=v_{c}\prod_{j=1}^{c-1}(1-v_{j})\quad\in[0,1]
\end{equation}
 and
\begin{equation}
\sum_{c=1}^{\infty}\varpi_{c}(\boldsymbol{v})=1
\end{equation}
Under the stick-breaking representation of the Pitman-Yor process,
the atoms $\Theta_{c}$, drawn independently from the base distribution
$G_{0}$, can be seen as the parameters of the component distributions
of a mixture model comprising an unbounded number of component densities,
with mixing proportions $\varpi_{c}(\boldsymbol{v})$.

\subsection{Gaussian process models}

Let us consider an observation space $\mathcal{X}$. A Gaussian process
$f(\boldsymbol{x}),\;\boldsymbol{x}\in\mathcal{X}$, is defined as
\emph{a collection of random variables, any finite number of which
have a joint Gaussian distribution} \cite{book}. A Gaussian process
is completely specified by its mean function and covariance function.
We define the mean function $m(\boldsymbol{x})$ and the covariance
function $k(\boldsymbol{x},\boldsymbol{x}')$ of a real process $f(\boldsymbol{x})$
as
\begin{equation}
\begin{aligned}m(\boldsymbol{x})= & \mathbb{E}[f(\boldsymbol{x})]\\
k(\boldsymbol{x},\boldsymbol{x}')= & \mathbb{E}[(f(\boldsymbol{x})-m(\boldsymbol{x}))(f(\boldsymbol{x}')-m(\boldsymbol{x}'))]
\end{aligned}
\end{equation}
 and we will write the Gaussian process as
\begin{equation}
f(\boldsymbol{x})\sim\mathcal{N}(m(\boldsymbol{x}),k(\boldsymbol{x},\boldsymbol{x}))
\end{equation}
Usually, for notational simplicity, and without any loss of generality,
the mean of the process is taken to be zero, $m(\boldsymbol{x})=0$,
although this is not necessary. Concerning selection of the covariance
function, a large variety of kernel functions $k(\boldsymbol{x},\boldsymbol{x}')$
might be employed, depending on the application considered \cite{book}.
This way, a postulated Gaussian process eventually takes the form
\begin{equation}
f(\boldsymbol{x})\sim\mathcal{N}(0,k(\boldsymbol{x},\boldsymbol{x})).
\end{equation}

Let us suppose a set of independent and identically distributed (i.i.d.)
samples $\mathcal{D}=\{(\boldsymbol{x}_{i},y_{i})|i=1,...,N\}$, with
the $d$-dimensional variables $\boldsymbol{x}_{i}$ being the observations
related to a modeled phenomenon, and the scalars $y_{i}$ being the
associated target values. The goal of a regression model is, given
a new observation $\boldsymbol{x}_{*}$, to predict the corresponding
target value $y_{*}$, based on the information contained in the training
set $\mathcal{D}$. The basic notion behind Gaussian process regression
consists in the assumption that the observable (training) target values
$y$ in a considered regression problem can be expressed as the superposition
of a Gaussian process over the input space $\mathcal{X}$, $f(\boldsymbol{x})$,
and an independent white Gaussian noise
\begin{equation}
y=f(\boldsymbol{x})+\epsilon
\end{equation}
 where $f(\boldsymbol{x})$ is given by (12), and
\begin{equation}
\epsilon\sim\mathcal{N}(0,\sigma^{2})
\end{equation}
Under this regard, the joint normality of the training target values
$\boldsymbol{y}=[y_{i}]_{i=1}^{N}$ and some unknown target value
$y_{\ast}$, approximated by the value $f_{\ast}$ of the postulated
Gaussian process evaluated at the observation point $\boldsymbol{x}_{\ast}$,
yields \cite{book}
\begin{equation}
\left[\begin{array}{c}
\boldsymbol{y}\\
f_{*}
\end{array}\right]\sim\mathcal{N}\left(\boldsymbol{0},\left[\begin{array}{cc}
\boldsymbol{K}(X,X)+\sigma^{2}\boldsymbol{I}_{N} & \boldsymbol{k}(\boldsymbol{x}_{*})\\
\boldsymbol{k}(\boldsymbol{x}_{*})^{T} & k(\boldsymbol{x}_{*},\boldsymbol{x}_{*})
\end{array}\right]\right)
\end{equation}
 where 
\begin{equation}
\boldsymbol{k}(\boldsymbol{x}_{*})\triangleq[k(\boldsymbol{x}_{1},\boldsymbol{x}_{*}),\dots,k(\boldsymbol{x}_{N},\boldsymbol{x}_{*})]^{T}
\end{equation}
 $X=\{\boldsymbol{x}_{i}\}_{i=1}^{N}$, $\boldsymbol{I}_{N}$ is the
$N\times N$ identity matrix, and $\boldsymbol{K}$ is the matrix
of the covariances between the $N$ training data points \emph{(design
matrix), }i.e.
\begin{equation}
\boldsymbol{K}(X,X)\triangleq\left[\begin{array}{ccc}
k(\boldsymbol{x}_{1},\boldsymbol{x}_{1}) & k(\boldsymbol{x}_{1},\boldsymbol{x}_{2})\dots & k(\boldsymbol{x}_{1},\boldsymbol{x}_{N})\\
k(\boldsymbol{x}_{2},\boldsymbol{x}_{1}) & k(\boldsymbol{x}_{2},\boldsymbol{x}_{2})\dots & k(\boldsymbol{x}_{2},\boldsymbol{x}_{N})\\
\vdots & \vdots & \vdots\\
k(\boldsymbol{x}_{N},\boldsymbol{x}_{1}) & k(\boldsymbol{x}_{N},\boldsymbol{x}_{2})\dots & k(\boldsymbol{x}_{N},\boldsymbol{x}_{N})
\end{array}\right]
\end{equation}

Then, from (16), and conditioning on the available training samples,
we can derive the expression of the model predictive distribution,
yielding
\begin{equation}
p(f_{*}|\boldsymbol{x}_{*},\mathcal{D})=\mathcal{N}(f_{*}|\mu_{*},\sigma_{*}^{2})
\end{equation}
\begin{equation}
\mu_{*d}^{c}=\boldsymbol{k}(\boldsymbol{x}_{*})^{T}(\boldsymbol{K}(X,X)+\sigma^{2}\boldsymbol{I}_{N})^{-1}\boldsymbol{y}
\end{equation}
\begin{equation}
\sigma_{*}^{2}=\sigma^{2}-\boldsymbol{k}(\boldsymbol{x}_{*})^{T}\left(\boldsymbol{K}(X,X)+\sigma^{2}\boldsymbol{I}_{N}\right)^{-1}\boldsymbol{k}(\boldsymbol{x}_{*})+k(\boldsymbol{x}_{*},\boldsymbol{x}_{*})
\end{equation}

Regarding optimization of the hyperparameters of the employed covariance
function (kernel), say $\boldsymbol{\theta}$, and the noise variance
$\sigma^{2}$ of a GP model, this is usually conducted by type-II
maximum likelihood, that is by maximization of the model marginal
likelihood (evidence). Using (16), it is easy to show that the evidence
of the GP regression model yields
\begin{equation}
\begin{aligned}\mathrm{log}p(y|X;\boldsymbol{\theta},\sigma^{2})= & -\frac{N}{2}\mathrm{log}2\pi-\frac{1}{2}\mathrm{log}\big|\boldsymbol{K}(X,X)+\sigma^{2}\boldsymbol{I}_{N}\big|\\
 & -\frac{1}{2}\boldsymbol{y}^{T}\big(\boldsymbol{K}(X,X)+\sigma^{2}\boldsymbol{I}_{N}\big)^{-1}\boldsymbol{y}
\end{aligned}
\end{equation}

It is interesting to note that the GP regression model considers that
the noise that contaminates the modeled output variables does not
depend on the observations themselves, but rather that it constitutes
an additive white noise term with constant variance, which bears no
correlation between observations, and no dependence on the values
of the observations. Nevertheless, in many real-world applications,
with financial return series modeling being a characteristic example,
this assumption of constant noise variance is clearly implausible.

To ameliorate this issue, an heteroscedastic GP regression approach
was proposed in \cite{tsitsias}, where the noise variance is considered
to be a function of the observed data, similar to previously proposed
\emph{heteroscedastic regression }approaches applied to econometrics
and statistical finance, e.g., \cite{kersting,brookes}. A key drawback
of the approach of \cite{tsitsias} is that their heteroscedastic
regression approach does not allow for capturing the clustering effects
in the variability or volatility, which is apparent in the vast majority
of financial return series data, and is effectively captured by GARCH-type
models. Our approach addresses these issues under a nonparametric
Bayesian mixture modeling scheme, as discussed next.

\section{Proposed Approach}

In this section, we first define the proposed MGPCH model, considering
a generic modeling problem that comprises the input variables $\boldsymbol{x}\in\mathbb{R}^{p}$,
and the output variables $\boldsymbol{y}\in\mathbb{R}^{D}$. Further,
we derive an efficient inference algorithm for our model under the
variational Bayesian inference paradigm, and we obtain the expression
of its predictive density. Finally, we show how we can obtain a predictive
distribution for the covariances between the modeled output variables
$\{y_{i}\}_{i=1}^{D}$, by utilization of the statistical tool of
\emph{copulas}.

\subsection{Model definition}

Let $f_{d}(\boldsymbol{x})$ be a\emph{ latent function} modeling
the $d$th output variable $y_{d}$ as a function of the model input
$\boldsymbol{x}$. We consider that the expression of $y_{d}$ as
a function of $\boldsymbol{x}$ is not uniquely described by the latent
function $f_{d}(\boldsymbol{x})$, but $f_{d}(\boldsymbol{x})$ is
only an instance of the (possibly infinite) set of possible latent
functions $f_{d}^{c}(\boldsymbol{x}),\; c=1,\dots,\infty$. To determine
the association between input samples and latent functions, we impose
a Pitman-Yor process prior over this set of functions. In addition,
we consider that each one of these latent functions $f_{d}^{c}(\boldsymbol{x})$
has a prior distribution of the form of a Gaussian process over the
whole space of input variables $\boldsymbol{x}\in\mathbb{R}^{p}$.
At this point, we make a\emph{ further key-assumption}: We assume
that the noise variance $\sigma^{2}$ of the postulated GPs is \emph{not}
a constant, but rather that \emph{it varies with the input} \emph{variables}
$\boldsymbol{x}\in\mathbb{R}^{p}$. In other words, we consider the
noise variance as a \emph{latent process}, different for each model
output variable, and exhibiting a clustering effect, as described
by the dynamics of the postulated PYP mixing prior.

Let us consider a set of input/output observation pairs $\{\boldsymbol{x}_{n},\boldsymbol{y}_{n}\}_{n=1}^{N}$,
comprising $N$ samples. Let us also introduce the set of variables
$\{z_{nc}\}_{n,c=1}^{N,\infty}$, with $z_{nc}=1$ if the function
relating $\boldsymbol{x}_{n}$ to $\boldsymbol{y}_{n}$ is considered
to be expressed by the set $\{f_{d}^{c}(\boldsymbol{x})\}_{d=1}^{D}$
of postulated Gaussian processes, $z_{nc}=0$ otherwise. Then, based
on the previous description, we essentially postulate the following
model:
\begin{equation}
p\big(\boldsymbol{y}_{n}|\boldsymbol{x}_{n},z_{nc}=1\big)=\prod_{d=1}^{D}\mathcal{N}(y_{nd}|f_{d}^{c}(\boldsymbol{x}_{n}),\sigma_{d}^{c}(\boldsymbol{x}_{n})^{2})
\end{equation}
\begin{equation}
p(z_{nc}=1|\boldsymbol{v})=\varpi_{c}(\boldsymbol{v})
\end{equation}
\begin{equation}
\varpi_{c}(\boldsymbol{v})=v_{c}\prod_{j=1}^{c-1}(1-v_{j})\in[0,1]
\end{equation}
 with 
\begin{equation}
\sum_{c=1}^{\infty}\varpi_{c}(\boldsymbol{v})=1\;
\end{equation}
\begin{equation}
p(v_{c})=\mathrm{Beta}(1-\delta,\alpha+\delta c)\;
\end{equation}
 and 
\begin{equation}
p(\boldsymbol{f}_{d}^{c}|X)=\mathcal{N}(\boldsymbol{f}_{d}^{c}|\boldsymbol{0},\boldsymbol{K}^{c}(X,X))
\end{equation}
where $y_{nd}$ is the $d$th element of $\boldsymbol{y}_{n}$, we
define $X\triangleq\{\boldsymbol{x}_{n}\}_{n=1}^{N}$, $Y\triangleq\{\boldsymbol{y}_{n}\}_{n=1}^{N}$,
and $Z\triangleq\{z_{nc}\}_{n,c=1}^{N,\infty}$, $\boldsymbol{f}_{d}^{c}$
is the vector of the $f_{d}^{c}(\boldsymbol{x}_{n})\;\forall n$,
i.e., $\boldsymbol{f}_{d}^{c}\triangleq[f_{d}^{c}(\boldsymbol{x}_{n})]_{n=1}^{N}$,
and $\boldsymbol{K}^{c}(X,X)$ is the following \emph{design matrix
}
\begin{equation}
\boldsymbol{K}^{c}(X,X)\triangleq\left[\begin{array}{ccc}
k^{c}(\boldsymbol{x}_{1},\boldsymbol{x}_{1}) & k^{c}(\boldsymbol{x}_{1},\boldsymbol{x}_{2})\dots & k^{c}(\boldsymbol{x}_{1},\boldsymbol{x}_{N})\\
k^{c}(\boldsymbol{x}_{2},\boldsymbol{x}_{1}) & k^{c}(\boldsymbol{x}_{2},\boldsymbol{x}_{2})\dots & k^{c}(\boldsymbol{x}_{2},\boldsymbol{x}_{N})\\
\vdots & \vdots & \vdots\\
k^{c}(\boldsymbol{x}_{N},\boldsymbol{x}_{1}) & k^{c}(\boldsymbol{x}_{N},\boldsymbol{x}_{2})\dots & k^{c}(\boldsymbol{x}_{N},\boldsymbol{x}_{N})
\end{array}\right]
\end{equation}

Regarding the latent processes $\sigma_{d}^{c}(\boldsymbol{x}_{n})^{2}$,
we choose to also impose a GP prior over them. Specifically, to accommodate
the fact that $\sigma_{d}^{c}(\boldsymbol{x}_{n})^{2}\geq0$ (by definition),
we postulate 
\begin{equation}
\sigma_{d}^{c}(\boldsymbol{x}_{n})^{2}=\mathrm{exp}\left[g_{d}^{c}(\boldsymbol{x}_{n})\right]
\end{equation}
 and
\begin{equation}
p\left(\boldsymbol{g}_{d}^{c}|X\right)=\mathcal{N}\left(\boldsymbol{g}_{d}^{c}|\tilde{m}_{d}^{c}\boldsymbol{1},\boldsymbol{\Lambda}^{c}(X,X)\right)\label{eq:p(gc|x;theta)}
\end{equation}
where $\boldsymbol{g}_{d}^{c}$ is the vector of the $g_{d}^{c}(\boldsymbol{x}_{n})\;\forall n$,
i.e., $\boldsymbol{g}_{d}^{c}\triangleq[g_{d}^{c}(\boldsymbol{x}_{n})]_{n=1}^{N}$,
and $\boldsymbol{\Lambda}^{c}(X,X)$ is a \emph{design matrix, }similar
to $\boldsymbol{K}^{c}(X,X)$, but with (possibly) different kernel
functions $\lambda(\cdot,\cdot)$. 

Finally, due to the effect of the innovation parameter $\alpha$ on
the number of effective mixture components, we also impose a Gamma
prior over it:
\begin{equation}
p(\alpha)=\mathcal{G}(\alpha|\eta_{1},\eta_{2}).
\end{equation}
 This completes the definition of our proposed MGPCH model.

\subsection{Inference algorithm}

Inference for nonparametric models can be conducted under a Bayesian
setting, typically by means of variational Bayes (e.g., \cite{vbdpm}),
or Monte Carlo techniques (e.g., \cite{dpm-hmm}). Here, we prefer
a variational Bayesian approach, due to its considerably better scalability
in terms of computational costs, which becomes of major importance
when having to deal with large data corpora \cite{tnn2,tnn3}. 

Our variational Bayesian inference algorithm for the MGPCH model comprises
derivation of a family of variational posterior distributions $q(.)$
which approximate the true posterior distribution over the infinite
sets $Z$, $\boldsymbol{v}={(v_{c})}_{c=1}^{\infty}$, $\{\boldsymbol{f}^{c}\}_{c=1}^{\infty}$,
and $\{\boldsymbol{g}^{c}\}_{c=1}^{\infty}$, and the innovation parameter
$\alpha$. Apparently, Bayesian inference is not tractable under this
setting, since we are dealing with an infinite number of parameters. 

For this reason, we employ a common strategy in the literature of
Bayesian nonparametrics, formulated on the basis of a truncated stick-breaking
representation of the PYP \cite{vbdpm}. That is, we fix a value $C$
and we let the variational posterior over the $v_{i}$ have the property
$q(v_{C}=1)=1$. In other words, we set $\varpi_{c}(\boldsymbol{v})$
equal to zero for $c>C$. Note that, under this setting, the treated
MGPCH model involves a full PYP prior; truncation is not imposed on
the model itself, but only on the variational distribution to allow
for tractable inference. Hence, the truncation level $C$ is a variational
parameter which can be freely set, and not part of the prior model
specification.

Let $W\triangleq\{\boldsymbol{v},\alpha,Z,\{\boldsymbol{f}^{c}\}_{c=1}^{C},\{\boldsymbol{g}^{c}\}_{c=1}^{C}\}$
be the set of all the parameters of the MGPCH model over which a prior
distribution has been imposed, and $\Xi$ be the set of the hyperparameters
of the model priors and kernel functions. Variational Bayesian inference
introduces an arbitrary distribution $q(W)$ to approximate the actual
posterior $p(W|\Xi,X,Y)$ which is computationally intractable, yielding
\cite{vbg}
\begin{equation}
\mathrm{log}p(X,Y)=\mathcal{L}(q)+\mathrm{KL}(q||p)
\end{equation}
where
\begin{equation}
\mathcal{L}(q)=\int\mathrm{d}Wq(W)\mathrm{log}\frac{p(X,Y,W|\Xi)}{q(W)}
\end{equation}
and $\mathrm{KL}(q||p)$ stands for the Kullback-Leibler (KL) divergence
between the (approximate) variational posterior, $q(W)$, and the
actual posterior, $p(W|\Xi,X,Y)$. Since KL divergence is nonnegative,
$\mathcal{L}(q)$ forms a strict lower bound of the log evidence,
and would become exact if $q(W)=p(W|\Xi,X,Y)$. Hence, by maximizing
this lower bound $\mathcal{L}(q)$ (variational free energy) so that
it becomes as tight as possible, not only do we minimize the KL-divergence
between the true and the variational posterior, but we also implicitly
integrate out the unknowns $W$.

Due to the considered conjugate exponential prior configuration of
the MGPCH model, the variational posterior $q(W)$ is expected to
take the same functional form as the prior, $p(W)$ \cite{vbtmfa}:
\begin{equation}
\begin{split}q(W)= & q(Z)q(\alpha)\left(\prod_{c=1}^{C-1}q(v_{c})\right)\prod_{c=1}^{C}\prod_{d=1}^{D}q\left(\boldsymbol{f}_{d}^{c}\right)q\left(\boldsymbol{g}_{d}^{c}\right)\end{split}
\end{equation}
 with
\begin{equation}
q(Z)=\prod_{n=1}^{N}\prod_{c=1}^{C}q(z_{nc}=1)
\end{equation}
 Then, the variational free energy of the model reads (ignoring constant
terms)

\begin{equation}
\begin{split}\mathcal{L}(q) & =\sum_{c=1}^{C}\sum_{d=1}^{D}\int\mathrm{d}\boldsymbol{f}_{d}^{c}q(\boldsymbol{f}_{d}^{c})\mathrm{log}\frac{p(\boldsymbol{f}_{d}^{c}|\boldsymbol{0},\boldsymbol{K}^{c}(X,X))}{q(\boldsymbol{f}_{d}^{c})}\\
+ & \sum_{c=1}^{C}\sum_{d=1}^{D}\int\mathrm{d}\boldsymbol{g}_{d}^{c}q(\boldsymbol{g}_{d}^{c})\mathrm{log}\frac{p(\boldsymbol{g}_{d}^{c}|\tilde{m}_{d}^{c}\boldsymbol{1},\boldsymbol{\Lambda}^{c}(X,X))}{q(\boldsymbol{g}_{d}^{c})}\\
+ & \int\mathrm{d}\alpha q(\alpha)\bigg\{\mathrm{log}\frac{p(\alpha|\eta_{1},\eta_{2})}{q(\alpha)}\\
 & +\sum_{c=1}^{C-1}\int\mathrm{d}v_{c}q(v_{c})\mathrm{log}\frac{p(v_{c}|\alpha)}{q(v_{c})}\bigg\}\\
+ & \sum_{c=1}^{C}\sum_{n=1}^{N}q(z_{nc}=1)\bigg\{\int\mathrm{d}\boldsymbol{v}q(\boldsymbol{v})\mathrm{log}\frac{p(z_{nc}=1|\boldsymbol{v})}{q(z_{nc}=1)}\\
+ & \bigg.\sum_{d=1}^{D}\int\int\mathrm{d}\boldsymbol{f}_{d}^{c}\mathrm{d}\boldsymbol{g}_{d}^{c}q(\boldsymbol{f}_{d}^{c})q(\boldsymbol{g}_{d}^{c})\mathrm{log}p(y_{nd}|f_{d}^{c}(\boldsymbol{x}_{n}),\sigma_{d}^{c}(\boldsymbol{x}_{n})^{2})\bigg\}
\end{split}
\end{equation}

Derivation of the variational posterior distribution $q(W)$ involves
maximization of the variational free energy $\mathcal{L}(q)$ over
each one of the factors of $q(W)$ in turn, holding the others fixed,
in an iterative manner \cite{statmech}. By construction, this iterative,
consecutive updating of the variational posterior distribution is
guaranteed to monotonically and maximally increase the free energy
$\mathcal{L}(q)$ \cite{vbtmfa}. 

Let us denote as $\left<.\right>$ the posterior expectation of a
quantity. From (37), we obtain the following variational (approximate)
posteriors over the parameters of our model:\\
1. Regarding the PYP stick-breaking variables $v_{c}$, we have
\begin{equation}
q(v_{c})=\mathrm{Beta}(v_{c}|\beta_{c,1},\beta_{c,2})
\end{equation}
 where
\begin{equation}
\beta_{c,1}=1-\delta+\sum_{n=1}^{N}q(z_{nc}=1)
\end{equation}
\begin{equation}
\beta_{c,2}=\left<\alpha\right>+c\delta+\sum_{c'=c+1}^{C}\sum_{n=1}^{N}q(z_{nc'}=1)
\end{equation}
 2. The innovation parameter $\alpha$ yields
\begin{equation}
q(\alpha)=\mathcal{G}(\alpha|\hat{\eta}_{1},\hat{\eta}_{2})
\end{equation}
 where
\begin{equation}
\hat{\eta}_{1}=\eta_{1}+C-1
\end{equation}
\begin{equation}
\hat{\eta}_{2}=\eta_{2}-\sum_{c=1}^{C-1}\left[\psi(\beta_{c,2})-\psi(\beta_{c,1}+\beta_{c,2})\right]
\end{equation}
 $\psi(.)$ denotes the Digamma function, and
\begin{equation}
\left<\alpha\right>=\frac{\hat{\eta}_{1}}{\hat{\eta}_{2}}
\end{equation}
 3. Regarding the posteriors over the latent functions $\boldsymbol{f}_{d}^{c}$,
we have
\begin{equation}
q(\boldsymbol{f}_{d}^{c})=\mathcal{N}(\boldsymbol{f}_{d}^{c}|\boldsymbol{\mu}_{d}^{c},\boldsymbol{\Sigma}_{d}^{c})
\end{equation}
 where 
\begin{equation}
\boldsymbol{\Sigma}_{d}^{c}=\left(\left(\boldsymbol{K}^{c}(X,X)\right)^{-1}+\boldsymbol{B}_{d}^{c}\right)^{-1}
\end{equation}
\begin{equation}
\boldsymbol{\mu}_{d}^{c}=\boldsymbol{\Sigma}_{d}^{c}\boldsymbol{B}_{d}^{c}\boldsymbol{y}_{d}
\end{equation}
\begin{equation}
\boldsymbol{B}_{d}^{c}\triangleq\mathrm{diag}\left(\left[\frac{1}{\left<\sigma_{d}^{c}(\boldsymbol{x}_{n})^{2}\right>}q(z_{nc}=1)\right]_{n=1}^{N}\right)
\end{equation}
 and $\boldsymbol{y}_{d}\triangleq[y_{nd}]_{n=1}^{N}$.\\
4. Similar, regarding the posteriors over the latent noise variance
processes $\boldsymbol{g}_{d}^{c}$, we have
\begin{equation}
q(\boldsymbol{g}_{d}^{c})=\mathcal{N}(\boldsymbol{g}_{d}^{c}|\boldsymbol{m}_{d}^{c},\boldsymbol{S}_{d}^{c})
\end{equation}
 where
\begin{equation}
\boldsymbol{S}_{d}^{c}=\left(\left(\boldsymbol{\Lambda}^{c}(X,X)\right)^{-1}+\boldsymbol{Q}_{d}^{c}\right)^{-1}
\end{equation}
\begin{equation}
\boldsymbol{m}_{d}^{c}=\boldsymbol{\Lambda}^{c}(X,X)\left(\boldsymbol{Q}_{d}^{c}-\frac{1}{2}\mathrm{diag}\left[q\left(z_{nc}=1\right)\right]_{n=1}^{N}\right)\boldsymbol{1}+\tilde{m}_{d}^{c}\boldsymbol{1}
\end{equation}
and $\boldsymbol{Q}_{d}^{c}$ is a positive semi-definite diagonal
matrix, whose components comprise variational parameters that can
be freely set. Note that, from this result, it follows 
\begin{equation}
\left<\sigma_{d}^{c}(\boldsymbol{x}_{n})^{2}\right>=\mathrm{exp}\left([\boldsymbol{m}_{d}^{c}]_{n}-\frac{1}{2}[\boldsymbol{S}_{d}^{c}]_{nn}\right)
\end{equation}
5. Finally, the posteriors over the latent variables $Z$ yield
\begin{equation}
q(z_{nc}=1)\propto\mathrm{exp}\left(\left<\mathrm{log}\varpi_{c}(\boldsymbol{v})\right>\right)\mathrm{exp}\left(r_{nc}\right)
\end{equation}
 where
\begin{equation}
\left<\mathrm{log}\varpi_{c}(\boldsymbol{v})\right>=\sum_{c'=1}^{c-1}\left<\mathrm{log}(1-v_{c'})\right>+\left<\mathrm{log}v_{c}\right>
\end{equation}
 and
\begin{equation}
\begin{aligned}r_{nc}\triangleq-\frac{1}{2}\sum_{d=1}^{D}\bigg\{ & \frac{1}{\left<\sigma_{d}^{c}(\boldsymbol{x}_{n})^{2}\right>}\left[\left(\boldsymbol{y}_{nd}-[\boldsymbol{\mu}_{d}^{c}]_{n}\right)^{2}+[\boldsymbol{\Sigma}_{d}^{c}]_{nn}\right]\\
 & +[\boldsymbol{m}_{d}^{c}]_{n}\bigg\}
\end{aligned}
\end{equation}
 where $[\boldsymbol{\xi}]_{n}$ is the $n$th element of vector $\boldsymbol{\xi}$,
$[\boldsymbol{\Sigma}_{d}^{c}]_{nn}$ is the $(n,n)$th element of
$\boldsymbol{\Sigma}_{d}^{c}$, and it holds
\begin{equation}
\left<\mathrm{log}v_{c}\right>=\psi(\beta_{c,1})-\psi(\beta_{c,1}+\beta_{c,2})
\end{equation}
\begin{equation}
\left<\mathrm{log}(1-v_{c})\right>=\psi(\beta_{c,2})-\psi(\beta_{c,1}+\beta_{c,2})
\end{equation}

As a final note, estimates of the values of the model hyperparameters
set $\Xi$, which comprises the hyperparameters of the model priors
and the kernel functions $k(\cdot,\cdot)$ and $\lambda(\cdot,\cdot)$,
are obtained by maximization of the model variational free energy
$\mathcal{L}(q)$ over each one of them. For this purpose, in this
paper we resort to utilization of the limited memory variant of the
BFGS algorithm (L-BFGS) \cite{l-bfgs}.

\subsection{Predictive density}

Let us consider the predictive distribution of the $d$th model output
variable corresponding to $\boldsymbol{x}_{*}$. To obtain it, we
begin by deriving the predictive posterior distribution over the latent
variables $\boldsymbol{f}$. Following the relevant derivations of
Section 2.2, we have
\begin{equation}
q\left(\boldsymbol{f_{*}}\right)=\sum_{c=1}^{C}\left\langle \varpi_{c}\left(\boldsymbol{v}\right)\right\rangle \prod_{d=1}^{D}\mathcal{N}\left(f_{*d}^{c}|a_{*d}^{c},\left(\sigma_{*d}^{c}\right)^{2}\right)
\end{equation}
 where
\begin{equation}
a_{*d}^{c}=\boldsymbol{k}^{c}(\boldsymbol{x}_{*})^{T}\left(\boldsymbol{K}^{c}(X,X)+\left(\boldsymbol{B}_{d}^{c}\right)^{-1}\right)^{-1}\boldsymbol{y}_{d}
\end{equation}
\begin{equation}
\begin{aligned}\left(\sigma_{*d}^{c}\right)^{2}=-\boldsymbol{k}^{c}(\boldsymbol{x}_{*})^{T}\left(\boldsymbol{K}^{c}(X,X)+\left(\boldsymbol{B}_{d}^{c}\right)^{-1}\right)^{-1}\boldsymbol{k}^{c}(\boldsymbol{x}_{*})\\
+k^{c}(\boldsymbol{x}_{*},\boldsymbol{x}_{*})
\end{aligned}
\end{equation}
\begin{equation}
\left<\varpi_{c}(\boldsymbol{v})\right>=\left<v_{c}\right>\prod_{j=1}^{c-1}\left(1-\left<v_{j}\right>\right)
\end{equation}
\begin{equation}
\left<v_{c}\right>=\frac{\beta_{c,1}}{\beta_{c,1}+\beta_{c,2}}
\end{equation}
 and 
\begin{equation}
\boldsymbol{k}(\boldsymbol{x}_{*})\triangleq[k(\boldsymbol{x}_{1},\boldsymbol{x}_{*}),...,k(\boldsymbol{x}_{N},\boldsymbol{x}_{*})]^{T}
\end{equation}

Further, we proceed to the predictive posterior distribution over
the latent variables $\boldsymbol{g}$; we yield
\begin{equation}
q\left(g_{*d}^{c}\right)=\mathcal{N}\left(g_{*d}^{c}|\tau_{*d}^{c},\varphi_{*d}^{c}\right)
\end{equation}
where
\begin{equation}
\tau_{*d}^{c}=\boldsymbol{\lambda}^{c}(\boldsymbol{x}_{*})^{T}\left(\boldsymbol{Q}_{d}^{c}-\frac{1}{2}\right)\boldsymbol{1}+\tilde{m}_{d}^{c}
\end{equation}

\begin{equation}
\varphi_{*d}^{c}=\lambda^{c}(\boldsymbol{x}_{*},\boldsymbol{x}_{*})^{T}-\boldsymbol{\lambda}^{c}(\boldsymbol{x}_{*})^{T}\left(\boldsymbol{\Lambda}_{d}^{c}+(\boldsymbol{Q}_{d}^{c})^{-1}\right)^{-1}\boldsymbol{\lambda}^{c}(\boldsymbol{x}_{*})
\end{equation}
 and
\begin{equation}
\boldsymbol{\lambda}(\boldsymbol{x}_{*})\triangleq[\lambda(\boldsymbol{x}_{1},\boldsymbol{x}_{*}),...,\lambda(\boldsymbol{x}_{N},\boldsymbol{x}_{*})]^{T}
\end{equation}

Based on these results, the predictive posterior of our model output
variables yields 
\begin{equation}
\begin{aligned}q(y_{*d}) & =\int\mathcal{N}\bigg(y_{*d}\bigg|\sum_{c=1}^{C}\left\langle \varpi_{c}\left(\boldsymbol{v}\right)\right\rangle a_{*d}^{c},\\
 & \sum_{c=1}^{C}\left\langle \pi_{c}\left(\boldsymbol{v}\right)\right\rangle ^{2}\left[\left(\sigma_{*d}^{c}\right)^{2}+\mathrm{exp}\left(g_{*d}^{c}\right)\right]\bigg)\mathrm{d}g_{*d}^{c}
\end{aligned}
\end{equation}
We note that this expression does not yield a Gaussian predictive
posterior. However, it is rather straightforward to compute the predictive
means and variances of $y_{*d}$. It holds
\begin{equation}
\hat{y}_{*d}=\mathbb{E}\big[y_{*d}|\boldsymbol{x_{*}};\mathcal{D}\big]=\sum_{c=1}^{C}\left\langle \varpi_{c}\left(\boldsymbol{v}\right)\right\rangle a_{*d}^{c}
\end{equation}
and
\begin{equation}
\mathbb{V}\left[y_{*d}|\boldsymbol{x}_{*};\mathcal{D}\right]=\sum_{c=1}^{C}\left\langle \pi_{c}\left(\boldsymbol{v}\right)\right\rangle ^{2}\left[\left(\sigma_{*d}^{c}\right)^{2}+\psi_{*d}^{c}\right]
\end{equation}
where
\begin{equation}
\psi_{*d}^{c}\triangleq\mathbb{E}[\mathrm{exp}(g_{*d}^{c})|\boldsymbol{x}_{*};\mathcal{D}]=\mathrm{exp}\left(\tau_{*d}^{c}+\frac{1}{2}\varphi_{*d}^{c}\right)
\end{equation}

\subsection{Learning the covariances between the modeled output variables}

As one can observe from (23), a characteristic of our proposed MGPCH
model is its assumption that the distribution of the modeled output
vectors $\boldsymbol{y}\in\mathbb{R}^{D}$ factorizes over their component
variables $\{y_{d}\}_{d=1}^{D}$. Indeed, this type of modeling is
largely the norm in Gaussian process-based modeling approaches \cite{book}.
This construction in essence implies that, under our approach, the
modeled output variables are considered independent, i.e. their covariance
is always assumed to be zero. However, when jointly modeling the return
series of various assets, the modeled output variables (asset returns)
are rather strongly correlated, and it is desired to be capable of
predicting the values of their covariances for any given input value. 

Existing approaches for resolving these issues of GP-based models
are based on the introduction of an additional kernel-based modeling
mechanism that allows for capturing this latent covariance structure
\cite{boyle,multigp,multigp2,convolved,twin}. For example, in \cite{boyle}
the authors propose utilization of a convolution process to induce
correlations between two output components. In \cite{convolved},
a generalization of the previous method is proposed for the case of
more than two modeled outputs combined under a convolved kernel. Along
the same lines, multitask learning approaches for resolving these
issues are presented in \cite{multigp} and \cite{multigp2}, where
separate GPs are postulated for each output, and are considered to
share the same prior in the context of a multitask learning framework. 

A drawback of the aforementioned existing approaches is that, in all
cases, learning entails employing a tedious optimization procedure
to estimate a large number of hyperparameters of the used kernel functions.
As expected, such a procedure is, indeed, highly prone to getting
trapped to bad local optima, a fact that might severely undermine
model performance. 

In this work, to avoid being confronted with such optimization issues,
and inspired by the financial research literature, we devise a novel
way of capturing the interdependencies between the modeled output
variables $\{y_{d}\}_{d=1}^{D}$, expressed in the form of their covariances:
specifically, we use the statistical tool of \emph{copulas} \cite{sklar}.
The copula, introduced in the seminal work of Sklar \cite{sklar},
is a model of statistical dependence between random variables. A copula
is defined as a multivariate distribution with standard uniform marginal
distributions, or, alternatively, as a function (with some restrictions
mentioned for example in \cite{copula-book}) that maps values from
the unit hypercube to values in the unit interval.

\subsubsection{Copulas: An introduction }

Let $\boldsymbol{y}=[y_{d}]_{d=1}^{D}$ be a $D$-dimensional random
variable with joint cumulative distribution function (cdf) $F\left([y_{d}]_{d=1}^{D}\right)$,
and marginal cdf's $F_{d}(y_{d}),\; d=1,\dots,D$, respectively. Then,
according to Sklar's theorem, there exists a $D$-variate copula cdf
$C(\cdot,\dots,\cdot)$ on $[0,1]^{D}$ such that
\begin{equation}
F\left(y_{1},\dots,y_{D}\right)=C\left(F_{1}(y_{1}),\dots,F_{D}(y_{D})\right)
\end{equation}
for any $\boldsymbol{y}\in\mathbb{R}^{D}$. Additionally, if the marginals
$F_{d}(\cdot),\; d=1,\dots,D$, are continuous, then the $D$-variate
copula $C(\cdot,\dots,\cdot)$ satisfying (72) is unique. Conversely,
if $C(\cdot,\dots,\cdot)$ is a $D$-dimensional copula and $F_{i}(\cdot),\; i=1,\dots,D$,
are univariate cdf's, it holds
\begin{equation}
C\left(u_{1},\dots,u_{D}\right)=F\left(F_{1}^{-1}(u_{1}),\dots,F_{D}^{-1}(u_{D})\right)
\end{equation}
 where $F_{d}^{-1}(\cdot)$ denotes the inverse of the cdf of the
$d$th marginal distribution $F_{d}(\cdot)$, i.e. the quantile function
of the $d$th modeled variable $y_{d}$. 

It is easy to show that the corresponding probability density function
of the copula model, widely known as the \emph{copula density function},
is given by
\begin{equation}
\begin{aligned}c\left(u_{1},\dots,u_{D}\right) & =\frac{\partial^{D}}{\partial u_{1}\dots\partial u_{D}}C\left(u_{1},\dots,u_{D}\right)\\
 & =\frac{\partial^{D}}{\partial u_{1}\dots\partial u_{D}}F\left(F_{1}^{-1}(u_{1}),\dots,F_{D}^{-1}(u_{D})\right)\\
 & =\frac{p\left(F_{1}^{-1}(u_{1}),\dots,F_{D}^{-1}(u_{D})\right)}{\prod_{i=1}^{D}p_{i}\left(F_{i}^{-1}(u_{i})\right)}
\end{aligned}
\end{equation}
 where $p_{i}\left(\cdot\right)$ is the probability density function
of the $i$th component variable $y_{i}$.

Let us now assume a parametric class for the copula $C(\cdot,\dots,\cdot)$
and the marginal cdf's $F_{i}(\cdot),\; i=1,\dots,D$, respectively.
In particular, let $\zeta$ denote the (trainable) parameter (or set
of parameters) of the postulated copula. Then, the joint probability
density of the modeled variables $\boldsymbol{y}=[y_{i}]_{i=1}^{D}$
yields
\begin{equation}
\begin{aligned}p(y_{1},\dots,y_{D}|\zeta)\\
=\left[\prod_{i=1}^{D}p_{i}(y_{i})\right] & c\left(F_{1}(y_{1}),\dots,F_{D}(y_{D})|\zeta\right)
\end{aligned}
\end{equation}

Since the emergence of the concept of copula, several copula families
have been constructed, e.g., Gaussian, Clayton, Frank, Gumbel, Joe,
etc, that enable capturing of any form of dependence structure. By
coupling different marginal distributions with different copula functions,
copula-based models are able to model a wide variety of marginal behaviors
(such as skewness and fat tails), and dependence properties (such
as clusters, positive or negative tail dependence) \cite{copula-book}.
Selection of the best-fit copula has been a topic of rigorous research
efforts during the last years, and motivating results have already
been achieved \cite{markov-copula} (for excellent and detailed discussions
on copulas, c.f. \cite{copula-book,joe}).

\subsubsection{Proposed Approach}

In this work, to capture the interdependencies (covariances) between
the MGPCH-modeled output variables, we propose a \emph{conditional
copula}-based dependence modeling framework. Specifically, for the
considered $D$-dimensional output vectors $\boldsymbol{y}=[y_{d}]_{d=1}^{D}$,
we postulate \emph{pairwise parametric conditional models }for each
output pair $(y_{i},y_{j})_{i,j=1,i\neq j}^{D}$, with cdf's defined
as follows:
\begin{equation}
F(y_{i},y_{j}|\boldsymbol{x})=C(F_{i}(y_{i}|\boldsymbol{x}),F_{j}(y_{j}|\boldsymbol{x})|\boldsymbol{x})
\end{equation}
where the marginals $F_{d}(y_{d}|\boldsymbol{x})$ are the cdf's that
correspond to the predictive posteriors $q(y_{*d})$ given by (68),
and the used input-conditional copulas are defined under a parametric
construction as
\begin{equation}
C(u_{i},u_{j}|\boldsymbol{x})\triangleq C(u_{i},u_{j}|\zeta_{ij}(\boldsymbol{x}))
\end{equation}
 and we consider that the $\zeta_{ij}(\boldsymbol{x})$ are given
by
\begin{equation}
\zeta_{ij}(\boldsymbol{x})=\xi(\gamma_{ij}(\boldsymbol{x}))
\end{equation}
where the $\gamma_{ij}(\boldsymbol{x})$ are trainable real-valued
models, and $\xi(\cdot)$ is a link function ensuring that the values
of $\zeta_{ij}(\boldsymbol{x})$ will always be within the range allowed
by the copula model employed each time. For instance, if a Clayton
copula $C(\cdot)$ is employed, it is required that its parameter
be positive, i.e. $\zeta_{ij}(\boldsymbol{x})>0$ \cite{copula-book};
in such a case, $\xi(\cdot)$ may be defined as the exponential function,
i.e. $\xi(\alpha)=\mathrm{exp}(\alpha)$. 

Note that the predictive posteriors $q(y_{*d})$ are difficult to
compute analytically, since (68) does not yield a Gaussian distribution.
For this reason, and in order to facilitate efficient training of
the postulated pairwise conditional copula models, in the following
we approximate (68) as a Gaussian with mean and variance given by
(69) and (70), respectively.

Further, we consider the functions $\gamma_{ij}(\boldsymbol{x})$
to be linear basis functions models. Specifically, we postulate
\begin{equation}
\gamma_{ij}(\boldsymbol{x})=\boldsymbol{w}_{ij}^{T}\boldsymbol{h}(\boldsymbol{x})
\end{equation}
where the $\boldsymbol{w}_{ij}$ are trainable model parameters, and
the basis functions $\boldsymbol{h}(\boldsymbol{x})$ are defined
using a small set of basis input observations $\{\boldsymbol{x}_{i}\}_{i=1}^{I}$,
and an appropriate kernel function $\tilde{k}$:
\begin{equation}
\boldsymbol{h}(\boldsymbol{x})\triangleq[\tilde{k}(\boldsymbol{x},\boldsymbol{x}_{i})]_{i=1}^{I}
\end{equation}
Training for the postulated pairwise conditional copula models can
be performed by optimizing the logarithm of the copula density function
that corresponds to the parametric conditional model (77), given a
set of training data $\mathcal{D}=(\boldsymbol{x}_{n},\boldsymbol{y}_{n})_{n=1}^{N}$,
which yields
\begin{equation}
\mathcal{P}_{ij}=\sum_{n=1}^{N}\mathrm{log}\; c\left(F_{i}(y_{ni}|\boldsymbol{x}_{n}),F_{j}(y_{nj}|\boldsymbol{x}_{n})\big|\xi\left(\boldsymbol{w}_{ij}^{T}\boldsymbol{h}(\boldsymbol{x}_{n})\right)\right),
\end{equation}
 with respect to the parameter vectors $\boldsymbol{w}_{ij}$. To
effect this procedure, in this paper we resort to the L-BFGS algorithm
\cite{l-bfgs}.

\newcounter{tempequationcounter}\begin{figure*}[!t]   \normalsize   \setcounter{tempequationcounter}{\value{equation}} \addtocounter{tempequationcounter}{1}  \begin{IEEEeqnarray}{rCl} \setcounter{equation}{82} \mathbb{V}\left[y_{*i},y_{*j}|\boldsymbol{x}_{*};\mathcal{D}\right]=\int\int\left[C\left(F_{i}(\kappa|\boldsymbol{x}_{*}),F_{j}(\kappa'|\boldsymbol{x}_{*})\big|\xi\left(\boldsymbol{w}_{ij}^{T}\boldsymbol{h}(\boldsymbol{x}_{*})\right)\right)-F_{i}(\kappa|\boldsymbol{x}_{*})F_{j}(\kappa'|\boldsymbol{x}_{*})\right]\mathrm{d}\kappa\mathrm{d}\kappa' \end{IEEEeqnarray}   \setcounter{equation}{\value{tempequationcounter}}   \hrulefill   \vspace*{4pt} \end{figure*}

After training the postulated pairwise models $C(u_{i},u_{j}|\zeta_{ij}(\boldsymbol{x}))$
$\forall i\neq j$, computation of the predictive covariance $\mathbb{V}\left[y_{*i},y_{*j}|\boldsymbol{x}_{*};\mathcal{D}\right]$
between the $i$th and the $j$th model output given the input observation
$\boldsymbol{x}_{*}$ can be conducted using the corresponding conditional
copula model and marginal predictive densities. Specifically, from
Hoeffding’s lemma \cite{hoeffding,hoeffding2,hoeffding3}, we directly
obtain {[}Eq. (82){]}; this latter integral can be approximated by
means of numerical analysis methods.

\section{Experimental Evaluation}

In this section, we elaborate on the application of our MGPCH approach
to volatility modeling for financial return series data. We perform
an experimental evaluation of its performance in volatility modeling,
and examine how it compares to state-of-the-art competitors. We also
assess the efficacy of the proposed copula-based approach for learning
the predictive covariances between the modeled output variables of
the MGPCH model. 

For this purpose, we consider modeling the daily return series of
various financial indices, including currency exchange rates, global
large-cap equity indices, and Euribor rates. We note that, in this
work, asset return $r(t)$ is defined as the difference between the
logarithm of the prices $p(t)$ in two subsequent time points, i.e.,
$r(t)\triangleq\mathrm{log}p(t)-\mathrm{log}p(t-1)$. All our source
codes were developed in MATLAB R2012a.

\subsection{Volatility prediction using the MGPCH model }

In this set of experiments, we consider three application scenarios:
\begin{itemize}
\item In the first scenario, we model the return series pertaining to the
following \emph{currency exchange rates,} over the period December
31, 1979 to December 31, 1998 (daily closing prices):\\
1. (AUD) Australian Dollar / US \$ \\
2. (GBP) UK Pound / US \$ \\
3. (CAD) Canadian Dollar / US \$ \\
4. (DKK) Danish Krone / US \$ \\
5. (FRF) French Franc / US \$ \\
6. (DEM) German Mark / US \$ \\
7. (JPY) Japanese Yen / US \$ \\
8. (CHF) Swiss Franc / US \$.
\item In the second scenario, we model the return series pertaining to the
following \emph{global large-cap equity indices,} for the business
days over the period April 27, 1993 to July 14, 2003 (daily closing
prices):\\
1. (TSX) Canadian TSX Composite \\
2. (CAC) French CAC 40 \\
3. (DAX) German DAX \\
4. (NIK) Japanese Nikkei 225 \\
5. (FTSE) UK FTSE 100 \\
6. (SP) US S\&P 500.
\item Finally, in the third scenario, we model the return series pertaining
to the following seven \emph{global large-cap equity indices} and
\emph{Euribor rates,} for the business days over the period February
7, 2001 to April 24, 2006 (daily closing prices for the first 6 indices,
and annual percentage rate converted to daily effective yield for
the last index):\\
1. (TSX) Canadian TSX Composite \\
2. (CAC) French CAC 40 \\
3. (DAX) German DAX \\
4. (NIK) Japanese Nikkei 225 \\
5. (FTSE) UK FTSE 100 \\
6. (SP) US S\&P 500 \\
7. (EB3M) Three-month Euribor rate.
\end{itemize}
These series have become standard benchmarks for assessing the performance
of volatility prediction algorithms \cite{mcl,copulaproc,brooks}.

In all the considered scenarios, the proposed MGPCH model is trained
using as input data, $\boldsymbol{x}(t)$, vectors containing the
daily returns of all the assets considered in each scenario. The corresponding
training output data $\boldsymbol{y}(t)$ essentially comprise the
same series of input vectors shifted one-step ahead. In other words,
the output series are defined as $\boldsymbol{y}(t)\triangleq\boldsymbol{r}(t+1),\; t>0$,
and the input series as $\boldsymbol{x}(t)\triangleq\boldsymbol{r}(t),\; t<T$,
where $T$ is the total duration of the modeled return series, and
the vectors $\boldsymbol{r}(t)$ contain the return values of all
the considered indices at time $t$. 

In our experiments, we evaluate the MGPCH model using zero kernels
for the mean process, i.e. $k^{c}(\boldsymbol{x},\boldsymbol{x}')=0\;\forall c$;
this construction allows for our model to remain consistent with the
existing literature, where it is typically considered that the modeled
return series constitute a zero-mean noise-only process, i.e. $f_{d}^{c}(\boldsymbol{x})=0\;\forall d,c$.
Note though that our approach can seamlessly deal with learning the
mean process $f_{d}^{c}(\boldsymbol{x})$, if a model for its covariance
is available. Further, we consider autoregressive kernels of order
one for the noise variance process of the model, of the form
\begin{equation}
\lambda^{c}(\boldsymbol{x},\boldsymbol{x}')=\frac{\sigma_{0}^{2}}{(1-\phi^{2})}\phi^{||\boldsymbol{x}-\boldsymbol{x}'||}
\end{equation}
 where the $\phi$ and $\sigma_{0}^{2}$ are model hyperparameters,
estimated by means of free energy optimization (using the L-BFGS algorithm).

To obtain some comparative results, we also evaluate: (i) a common
baseline approach from the field of financial engineering and econometrics,
namely the GARCH(1,1) model \cite{bolerslev}, that is a GARCH model
with volatility terms of order one and residual terms of order one;
and (ii) the recently proposed VHGP approach of \cite{tsitsias}.
Both these approaches have been shown to be very competitive in the
task of volatility prediction in financial return series \cite{hansen,tsitsias}.
Note that the GARCH(1,1) model uses as input the time variable, while
the VHGP model is trained similar to MGPCH. 

In our experiments, similar to \cite{copulaproc}, all the evaluated
methods are trained using a rolling window of the previous 120 days
of returns to make 1, 7, and 30 days ahead volatility forecasts; we
retrain the models every 7 days. We use two performance metrics to
evaluate the considered algorithms: The first one is the mean squared
error (MSE) between the model-estimated volatilities and the squared
returns of the modeled return series. The second one is the MSE between
the generated predictions and the historical volatilities computed
over rolling windows of 10 contiguous return values (days). As discussed
in \cite{volatilityguide}, these two groundtruth measurements (squared
returns and historical volatilities) constitute two of the few consistent
ways of volatility measuring.

In Tables 1-3, we provide the obtained results for the three considered
scenarios. These results are means of the obtained MSEs over all the
assets modeled in each scenario. As we observe, our approach yields
a clear advantage and a significant improvement over its competitors,
of at least one order of magnitude, in all the considered scenarios,
in terms of both the employed evaluation metrics.

\begin{table*}
\caption{First Scenario: Obtained MSEs considering comparison against both
the squared returns and historical volatility.}

\centering{}%
\begin{tabular}{|c||c|c|c|c||c|c|c|c|}
\hline 
Evaluation Metric & \multicolumn{4}{c||}{Squared Returns} & \multicolumn{4}{c|}{Historical Volatility}\tabularnewline
\hline 
\hline 
Prediction Horizon & 1-step & 7-step & 30-step & Average & 1-step & 7-step & 30-step & Average\tabularnewline
\hline 
GARCH & 4.97$\times10^{-7}$ & 4.99$\times10^{-7}$ & 5.11$\times10^{-7}$ & 5.03$\times10^{-7}$ & 4.98$\times10^{-7}$ & 5.01$\times10^{-7}$ & 5.08$\times10^{-7}$ & 5.02$\times10^{-7}$\tabularnewline
\hline 
VHGP & 2.13$\times10^{-8}$ & 2.15$\times10^{-8}$ & 2.16$\times10^{-8}$ & 2.15$\times10^{-8}$ & 1.63$\times10^{-8}$ & 1.63$\times10^{-8}$ & 1.62$\times10^{-8}$ & 1.63$\times10^{-8}$\tabularnewline
\hline 
MGPCH & 1.46$\times10^{-8}$ & 1.46$\times10^{-8}$ & 1.48$\times10^{-8}$ & 1.47$\times10^{-8}$ & 1.03$\times10^{-9}$ & 1.02$\times10^{-9}$ & 1.02$\times10^{-9}$ & 1.03$\times10^{-9}$\tabularnewline
\hline 
\end{tabular}
\end{table*}

\begin{table*}
\caption{Second Scenario: Obtained MSEs considering comparison against both
the squared returns and historical volatility.}

\centering{}%
\begin{tabular}{|c||c|c|c|c||c|c|c|c|}
\hline 
Evaluation Metric & \multicolumn{4}{c||}{Squared Returns} & \multicolumn{4}{c|}{Historical Volatility}\tabularnewline
\hline 
\hline 
Prediction Horizon & 1-step & 7-step & 30-step & Average & 1-step & 7-step & 30-step & Average\tabularnewline
\hline 
GARCH & 9.47$\times10^{-6}$ & 9.56$\times10^{-6}$ & 9.96$\times10^{-6}$ & 9.66$\times10^{-6}$ & 9.99$\times10^{-6}$ & 1.00$\times10^{-5}$ & 1.03$\times10^{-5}$ & 1.01$\times10^{-5}$\tabularnewline
\hline 
VHGP & 3.40$\times10^{-7}$ & 3.42$\times10^{-7}$ & 3.53$\times10^{-7}$ & 3.45$\times10^{-7}$ & 4.10$\times10^{-7}$ & 4.06$\times10^{-7}$ & 3.98$\times10^{-7}$ & 4.05$\times10^{-7}$\tabularnewline
\hline 
MGPCH & 1.37$\times10^{-7}$ & 1.39$\times10^{-7}$ & 1.45$\times10^{-7}$ & 1.41$\times10^{-7}$ & 3.52$\times10^{-8}$ & 3.52$\times10^{-8}$ & 3.47$\times10^{-8}$ & 3.50$\times10^{-8}$\tabularnewline
\hline 
\end{tabular}
\end{table*}

\begin{table*}
\caption{Third Scenario: Obtained MSEs considering comparison against both
the squared returns and historical volatility.}

\centering{}%
\begin{tabular}{|c||c|c|c|c||c|c|c|c|}
\hline 
Evaluation Metric & \multicolumn{4}{c||}{Squared Returns} & \multicolumn{4}{c|}{Historical Volatility}\tabularnewline
\hline 
\hline 
Prediction Horizon & 1-step & 7-step & 30-step & Average & 1-step & 7-step & 30-step & Average\tabularnewline
\hline 
GARCH & 3.91$\times10^{-7}$ & 4.14$\times10^{-7}$ & 4.94$\times10^{-7}$ & 4.33$\times10^{-7}$ & 4.50$\times10^{-7}$ & 4.61$\times10^{-7}$ & 5.06$\times10^{-7}$ & 4.72$\times10^{-7}$\tabularnewline
\hline 
VHGP & 3.94$\times10^{-7}$ & 4.03$\times10^{-7}$ & 4.26$\times10^{-7}$ & 4.08$\times10^{-7}$ & 4.87$\times10^{-7}$ & 4.89$\times10^{-7}$ & 4.91$\times10^{-7}$ & 4.88$\times10^{-7}$\tabularnewline
\hline 
MGPCH & 1.44$\times10^{-7}$ & 1.45$\times10^{-7}$ & 1.52$\times10^{-7}$ & 1.47$\times10^{-7}$ & 4.36$\times10^{-8}$ & 4.42$\times10^{-8}$ & 4.42$\times10^{-8}$ & 4.40$\times10^{-8}$\tabularnewline
\hline 
\end{tabular}
\end{table*}

\subsection{Copula-based modeling of the covariances between asset returns}

Here, we evaluate the performance of the proposed copula-based approach
for learning a predictive model of the covariances between the MGPCH-modeled
asset returns. For this purpose, we repeat the previous experimental
scenarios, with the goal now being to obtain predictions regarding
the covariances between the assets modeled each time. 

In our experiments, we consider application of three popular Archimedean
copula types, namely \emph{Clayton, Frank, }and \emph{Gumbel }copulas
\cite{copula-book}. The employed MGPCH models are trained similar
to the previous experiments. The postulated conditional-copula pairwise
models use a basis set of input observations (to compute the $\boldsymbol{h}(\boldsymbol{x})$
in (80)) that comprises the 10\% of the available training data points,
i.e. 12 data points sampled at regular time intervals (one sample
every 10 days). 

To obtain some comparative results, we also evaluate the performance
of two state-of-the-art methods used for modeling dynamic covariance
matrices (multivariate volatility) for high-dimensional vector-valued
observations; specifically, we consider the CCC-MVGARCH(1,1) approach
of \cite{mv}, and the GARCH-BEKK(1,1) method of \cite{bekk}. As
our evaluation metric, we use the products of the returns of the corresponding
asset pairs at each time point. Our obtained results are depicted
in Tables 4-6. We observe that our approach yields a very competitive
result: specifically, in two out of the three considered scenarios,
the yielded improvement was equal to or exceeded one order of magnitude,
while, in one case, all methods yielded comparable results. We also
observe that switching the employed Archimedean copula type had only
marginal effects on model performance, in all our experiments. 

\begin{table*}
\caption{First Scenario: Obtained MSEs considering comparison against the asset
pair return products.}

\centering{}%
\begin{tabular}{|c||c|c|c|c|}
\hline 
Evaluation Metric & \multicolumn{4}{c|}{Squared Returns}\tabularnewline
\hline 
\hline 
Prediction Horizon & 1-step & 7-step & 30-step & Average\tabularnewline
\hline 
\hline 
CCC-MVGARCH & 1.19$\times10^{-7}$ & 1.18$\times10^{-7}$ & 1.20$\times10^{-7}$ & 1.20$\times10^{-7}$\tabularnewline
\hline 
BEKK & 1.16$\times10^{-7}$ & 1.16$\times10^{-7}$ & 1.17$\times10^{-7}$ & 1.16$\times10^{-7}$\tabularnewline
\hline 
\hline 
MGPCH: \emph{Clayton} & 1.16$\times10^{-7}$ & 1.17$\times10^{-7}$ & 1.17$\times10^{-7}$ & 1.17$\times10^{-7}$\tabularnewline
\hline 
MGPCH: \emph{Frank} & 1.16$\times10^{-7}$ & 1.16$\times10^{-7}$ & 1.17$\times10^{-7}$ & 1.17$\times10^{-7}$\tabularnewline
\hline 
MGPCH: \emph{Gumbel} & 1.16$\times10^{-7}$ & 1.16$\times10^{-7}$ & 1.16$\times10^{-7}$ & 1.16$\times10^{-7}$\tabularnewline
\hline 
\end{tabular}
\end{table*}

\begin{table*}
\caption{Second Scenario: Obtained MSEs considering comparison against the
asset pair return products.}

\centering{}%
\begin{tabular}{|c||c|c|c|c|}
\hline 
Evaluation Metric & \multicolumn{4}{c|}{Squared Returns}\tabularnewline
\hline 
\hline 
Prediction Horizon & 1-step & 7-step & 30-step & Average\tabularnewline
\hline 
\hline 
CCC-MVGARCH & 1.4$\times10^{-6}$ & 1.4$\times10^{-6}$ & 1.4$\times10^{-6}$ & 1.4$\times10^{-6}$\tabularnewline
\hline 
BEKK & 1.7$\times10^{-6}$ & 1.7$\times10^{-6}$ & 1.7$\times10^{-6}$ & 1.7$\times10^{-6}$\tabularnewline
\hline 
\hline 
MGPCH: \emph{Clayton} & 3.1$\times10^{-7}$ & 3.1$\times10^{-7}$ & 3.1$\times10^{-7}$ & 3.1$\times10^{-7}$\tabularnewline
\hline 
MGPCH: \emph{Frank} & 3.2$\times10^{-7}$ & 3.2$\times10^{-7}$ & 3.2$\times10^{-7}$ & 3.2$\times10^{-7}$\tabularnewline
\hline 
MGPCH: \emph{Gumbel} & 3.1$\times10^{-7}$ & 3.1$\times10^{-7}$ & 3.1$\times10^{-7}$ & 3.1$\times10^{-7}$\tabularnewline
\hline 
\end{tabular}
\end{table*}

\begin{table*}
\caption{Third Scenario: Obtained MSEs considering comparison against the asset
pair return products.}

\centering{}%
\begin{tabular}{|c||c|c|c|c|}
\hline 
Evaluation Metric & \multicolumn{4}{c|}{Squared Returns}\tabularnewline
\hline 
\hline 
Prediction Horizon & 1-step & 7-step & 30-step & Average\tabularnewline
\hline 
\hline 
CCC-MVGARCH & 0.0044 & 0.0045 & 0.0047 & 0.0045\tabularnewline
\hline 
BEKK & 0.84 & 0.85 & 0.85 & 0.85\tabularnewline
\hline 
\hline 
MGPCH: \emph{Clayton} & 9.81$\times10^{-5}$ & 9.81$\times10^{-5}$ & 9.85$\times10^{-5}$ & 9.83$\times10^{-5}$\tabularnewline
\hline 
MGPCH: \emph{Frank} & 9.82$\times10^{-5}$ & 9.81$\times10^{-5}$ & 9.82$\times10^{-5}$ & 9.82$\times10^{-5}$\tabularnewline
\hline 
MGPCH: \emph{Gumbel} & 9.81$\times10^{-5}$ & 9.82$\times10^{-5}$ & 9.82$\times10^{-5}$ & 9.82$\times10^{-5}$\tabularnewline
\hline 
\end{tabular}
\end{table*}

\section{Conclusions}

In this paper, we proposed a novel nonparametric Bayesian approach
for modeling conditional heteroscedasticity in financial return series.
Our approach consists in the postulation of a mixture of Gaussian
process regression models, each component of which models the noise
variance process that contaminates the observed data as a separate
latent Gaussian process driven by the observed data. We imposed a
nonparametric prior with power-law nature over the distribution of
the model mixture components, namely the Pitman-Yor process prior,
to allow for better capturing modeled data distributions with heavy
tails and skewness. In addition, in order to provide a predictive
posterior for the covariances over the modeled asset returns, we devised
a copula-based covariance modeling procedure built on top of our model.
To assess the efficacy of our approach, we applied it to several asset
return series, and compared its performance to several state-of-the-art
methods in the field, on the grounds of standard evaluation metrics.
As we observed, our approach yields a clear performance improvement
over its competitors in all the considered scenarios. 

\bibliographystyle{IEEEtran}


\end{document}